%% file: paper.tex
\documentclass[sigconf,usenames,dvipsnames]{acmart}

\AtBeginDocument{\providecommand\BibTeX{{\normalfont B\kern-0.5em{\scshape i\kern-0.25em b}\kern-0.8em\TeX}}}

\copyrightyear{2022}
\acmYear{2022}
\setcopyright{rightsretained}
\acmConference[WSDM '22] {Proceedings of the Fifteenth ACM International Conference on Web Search and Data Mining}{February 21--25, 2022}{Tempe, AZ, USA.}
\acmBooktitle{Proceedings of the Fifteenth ACM International Conference on Web Search and Data Mining (WSDM '22), February 21--25, 2022, Tempe, AZ, USA}
\acmPrice{}
\acmISBN{978-1-4503-9132-0/22/02}
\acmDOI{10.1145/3488560.3498451}

\usepackage[utf8]{inputenc}
\usepackage[T1]{fontenc}

\usepackage{url}
\usepackage{booktabs,cellspace,tabularx}
\usepackage{amsfonts}
\usepackage{nicefrac}
\usepackage{microtype}
\usepackage{xcolor}

\usepackage{colortbl}
\usepackage{amsmath}
\usepackage{amssymb}
\usepackage{bm}
\usepackage{graphicx}
\usepackage{paralist}
\usepackage{enumitem}
\setlist[itemize]{nosep,leftmargin=2em}
\setlist[enumerate]{nosep,leftmargin=2em}
\usepackage{xspace}
\usepackage{makecell}
\usepackage{cleveref}
\usepackage[vlined,ruled]{algorithm2e}
\usepackage{multirow}
\usepackage{caption}
\usepackage{subcaption}
\usepackage{wrapfig}
\usepackage{floatflt}
\usepackage{pdfrender}
\usepackage{lipsum}
\usepackage{kantlipsum}
\usepackage{kotex}
\usepackage{diagbox}
\usepackage{nicefrac}
\usepackage{balance}

\input{dfn}

\begin{document}
\fancyhead{}

\title{\method: Jointly Modeling Event Time and Network Structure for Reasoning over Temporal Knowledge Graphs}

\author{\fontsize{11.5pt}{12pt}\selectfont Namyong Park$^{1}$, Fuchen Liu$^2$, Purvanshi Mehta$^2$, Dana Cristofor$^2$, Christos Faloutsos$^{1}$, Yuxiao Dong$^2$}
\email{{namyongp,christos}@cs.cmu.edu,{fuchen.liu,purvmehta,danac}@microsoft.com,ericdongyx@gmail.com}
\affiliation{\institution{\textsuperscript{1}Carnegie Mellon University, \textsuperscript{2}Microsoft}
\country{}}

\renewcommand{\shortauthors}{N. Park et al.}

\pdfinfo{
/Author (Namyong Park, Fuchen Liu, Purvanshi Mehta, Dana Cristofor, Christos Faloutsos, Yuxiao Dong)
}

\begin{abstract}
How can we perform knowledge reasoning over temporal knowledge graphs (TKGs)?
TKGs represent facts about entities and their relations, 
where each fact is associated with a timestamp.
Reasoning over TKGs, i.e., inferring new facts from time-evolving KGs, is crucial for many applications to provide intelligent services.
However, despite the prevalence of real-world data that can be represented as TKGs, 
most methods focus on reasoning over static knowledge graphs, or cannot predict future events.
In this paper, we present a problem formulation that unifies 
the two major problems that need to be addressed for an effective reasoning over TKGs, 
namely, modeling the event time and the evolving network structure.
Our proposed method \method jointly models both tasks in an effective framework,
which captures the ever-changing structural and temporal dynamics in TKGs via recurrent event modeling, and
models the interactions between entities based on the temporal neighborhood aggregation framework.
Further, \method achieves an accurate modeling of event time, using flexible and efficient mechanisms based on neural density estimation.
Experiments show that \method outperforms existing methods in terms of effectiveness (up to \textit{77\%} and \textit{116\%} more accurate time and link prediction) and efficiency.
\vspace{-0.5em}
\end{abstract}

\ccsdesc[500]{Computing methodologies~Knowledge representation and reasoning}
\ccsdesc[500]{Computing methodologies~Temporal reasoning}
\ccsdesc[300]{Computing methodologies~Neural networks}

\keywords{temporal knowledge graphs, reasoning over temporal knowledge graphs, temporal point processes, 
graph representation learning, event time prediction, temporal link prediction}

\maketitle

\vspace{-0.5em}
{\fontsize{8pt}{8pt} \selectfont
\textbf{ACM Reference Format:}\\
Namyong Park, Fuchen Liu, Purvanshi Mehta, Dana Cristofor, Christos Faloutsos, Yuxiao Dong. 2022.
EvoKG: Jointly Modeling Event Time and Network Structure for Reasoning over Temporal Knowledge Graphs. In \textit{Proceedings of the Fifteenth ACM
International Conference on Web Search and Data Mining (WSDM ’22), February 21–25, 2022, Tempe, AZ, USA.} ACM, New York, NY, USA, 10	 pages.
https://doi.org/10.1145/3488560.3498451}

\vspace{-0.7em}
\section{Introduction}
\vspace{-0.3em}
\label{sec:intro}

\input{010introduction}

\section{Problem Formulation}
\label{sec:problem}
\input{020problem}

\section{Modeling a Temporal Knowledge Graph}
\label{sec:method}

\input{030method}

\section{Experiments}
\label{sec:exp}
\input{040experiments}

\vspace{-0.5em}
\section{Related Work}
\label{sec:related}
\input{050related}

\vspace{-0.7em}
\section{Conclusion}
\vspace{-0.3em}
\label{sec:concl}
\input{060conclusion}

\setcounter{table}{0}
\renewcommand{\thetable}{\Alph{section}\arabic{table}}
\appendix

\begin{acks}
This work was funded by Carnegie Mellon University CyLab, with generous support from Microsoft.
Namyong Park was supported by the Bloomberg Data Science Ph.D. Fellowship and the ILJU Foundation Ph.D. Fellowship.
\end{acks}

\clearpage

\bibliographystyle{ACM-Reference-Format}
\balance

\appendix

\clearpage
\section{Appendix}
\label{sec:appendix}

\input{070appendix}

\end{document}

%% file: dfn.tex
\newcommand\dunderline[3][-1pt]{{\sbox0{#3}\ooalign{\copy0\cr\rule[\dimexpr#1-#2\relax]{\wd0}{#2}}}}

\definecolor{Gray}{gray}{0.9}
\definecolor{LightGray}{gray}{0.94}
\definecolor{LightCyan}{rgb}{0.88,1,1}
\definecolor{Highlight}{rgb}{0.92,1,1}

\newcommand{\secondbest}[1]{\dunderline{0.8pt}{#1}}

\newcommand*{\belowrulesepcolor}[1]{\noalign{\kern-\belowrulesep
		\begingroup
		\color{#1}\hrule height\belowrulesep
		\endgroup
	}}
\newcommand*{\aboverulesepcolor}[1]{\noalign{\begingroup
		\color{#1}\hrule height\aboverulesep
		\endgroup
		\kern-\aboverulesep
	}}

\newcommand*{\boldcheckmark}{\textpdfrender{
		TextRenderingMode=FillStroke,
		LineWidth=.7pt, }{\checkmark}}

\newcommand{\cm}{\checkmark}

\newcommand{\overbar}[1]{\mkern 1.5mu\overline{\mkern-1.5mu#1\mkern-1.5mu}\mkern 1.5mu}

\newcommand{\vect}[1]{\bm{#1}}
\newcommand{\emb}[1]{\bm{\mathsf{#1}}}
\newcommand{\mat}[1]{\bm{\mathbf{\MakeUppercase{#1}}}}

\newcommand{\g}{G}
\newcommand{\priorG}{\g_{<t}}

\newcommand{\NA}{\textsc{n/a}}

\newcommand{\rnn}{\textsc{RNN}\xspace}
\newcommand{\mlp}{$ \mathrm{MLP} $\xspace}

\newcommand{\icews}{ICEWS18\xspace}
\newcommand{\icewsFourteen}{ICEWS14\xspace}
\newcommand{\icewsSmall}{ICEWS-500\xspace}
\newcommand{\gdelt}{GDELT\xspace}
\newcommand{\wiki}{WIKI\xspace}
\newcommand{\yago}{YAGO\xspace}

\newcommand{\method}{EvoKG\xspace}

\newcommand{\renet}{{RE-Net}\xspace}
\newcommand{\knowevolve}{{Know-Evolve}\xspace}
\newcommand{\dyrep}{{DyRep}\xspace}
\newcommand{\ghnn}{{GHNN}\xspace}
\newcommand{\litsee}{{LiTSEE}\xspace}
\newcommand{\rtpp}{{RTPP}\xspace}
\newcommand{\mhp}{{MHP}\xspace}

\newcommand{\codedataurl}{https://namyongpark.github.io/evokg}

%% file: 010introduction.tex
\begin{figure}[htbp!]
	\par\vspace{-0.5em}\par
	\centering
	\includegraphics[width=0.30\textwidth]{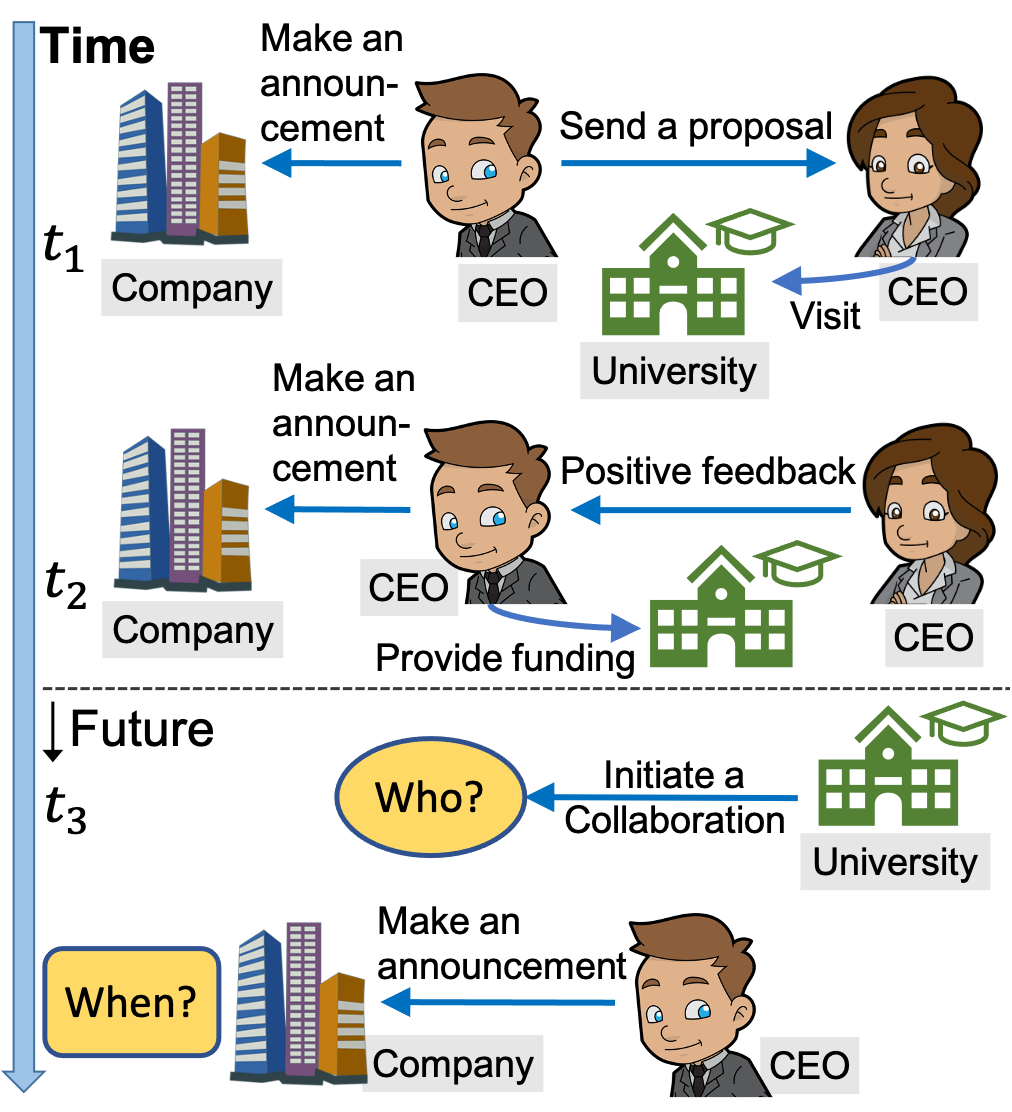}
	\setlength{\abovecaptionskip}{1pt}
\caption{An example TKG, where we aim to predict temporal links and event time.}
	\label{fig:tkg}
	\vspace{-2em}
\end{figure}

How can we perform knowledge reasoning over knowledge graphs (KGs) that continuously evolve over time?
KGs~\cite{DBLP:journals/corr/abs-2002-00388} organize and represent facts on various types of entities and their relations.
By facilitating an effective use of prior knowledge represented as a multi-relational graph,
KGs power many important applications, including question answering, recommender systems, search engines, and natural language processing.
Knowledge reasoning over KGs~\cite{DBLP:journals/eswa/ChenJX20}, the process of inferring new knowledge from existing facts in KGs, 
lies at the heart of these applications, 
as KGs are typically incomplete, with many facts missing.

Importantly, real-world events and facts are often associated with time (i.e., occurring at a specific time or valid in limited time), exhibiting complex dynamics among entities and their relations that evolve over time.
Such real-world data (e.g., ICEWS~\cite{boschee2015icews} and GDELT~\cite{leetaru2013gdelt}) can be modeled as temporal knowledge graphs (TKGs), 
where entities are connected via timestamped edges, and two entities can have multiple interactions at different time steps,
as illustrated in \Cref{fig:tkg}. Despite the prevalence of real-world data that can be represented as TKGs, 
existing methods~\cite{DBLP:journals/corr/YangYHGD14a,DBLP:conf/esws/SchlichtkrullKB18,
DBLP:conf/aaai/DettmersMS018,DBLP:conf/iclr/SunDNT19} have mainly focused on reasoning over static KGs, and 
lack the ability to employ rich temporal dynamics available in TKGs.

Recently, a few methods have been developed for reasoning over TKGs. 
They mainly address two problem setups, i.e., interpolation and extrapolation.
Given a TKG ranging from time~$ 0 $ to time~$ T $,
methods for the interpolation setup~\cite{DBLP:conf/emnlp/DasguptaRT18,DBLP:conf/emnlp/Garcia-DuranDN18,DBLP:conf/www/LeblayC18}
infer missing facts for time $ t ~ (0 \le t \le T) $;
on the other hand, those for the extrapolation setup~\cite{DBLP:conf/icml/TrivediDWS17,DBLP:conf/iclr/TrivediFBZ19,DBLP:conf/emnlp/JinQJR20}
predict new facts for time $ t > T $.
In this paper, we focus on the extrapolation setting, which is more challenging and interesting than the other setting, 
as forecasting emerging events are of great importance to many applications of TKG reasoning.

In this paper, we approach the problem of TKG modeling 
by defining the joint probability distribution of a TKG as a product of conditionals,
from which we present a problem formulation that unifies 
the two problem settings of existing methods, namely, modeling the event time and evolving network structure.
While addressing both problems leads to learning rich, complementary information useful for an effective reasoning over TKGs,
most methods deal with only either of the two, as summarized in \Cref{tab:salesmatrix}.

Therefore, in this work, we develop \method, a method that jointly addresses these two core tasks for reasoning over TKGs.
We design an effective framework that can be effectively applied to each task, with only minor adaptations.
Our framework performs neighborhood aggregation in a relation- and time-aware manner, and 
carries out recurrent event modeling in an autoregressive architecture
to capture the ever-changing structural and temporal dynamics over time (F1-F3 in \Cref{tab:salesmatrix}).
Importantly, \method tackles the challenging task of event time modeling,
using flexible and efficient mechanisms based on neural density estimation (T2-1 and T2-2 in \Cref{tab:salesmatrix}),
which avoids the limitations of existing methods that 
the learned distributions are not expressive, and 
that the log-likelihood and expectation of event time cannot be obtained in closed form, but instead require an approximation.
In summary, our contributions are as follows.
\begin{itemize}[nosep,leftmargin=1em]
\item \textbf{Problem Formulation} (\Cref{sec:problem}). We present a problem formulation 
that unifies the two major tasks for TKG reasoning---modeling the timing of events and evolving network structure.
\item \textbf{Framework} (\Cref{sec:method}). We propose \method, an effective and efficient method for reasoning over TKGs 
that jointly addresses the two core problems (T1 and T2 in \Cref{tab:salesmatrix}).
\item \textbf{Effectiveness} (\Cref{sec:exp}). Experiments show that \method achieves up to \textit{116\%} and \textit{77\%} better link and event time prediction accuracy, respectively, than existing KG reasoning methods (\Cref{fig:crownjewel}).
\item \textbf{Efficiency} (\Cref{sec:exp}). \method efficiently processes concurrent events, 
achieving up to \textit{30$ \times $} and \textit{291$ \times $} speedup in training and inference, respectively, compared to the best existing method.

\end{itemize}
\textbf{Reproducibility.} The code and data used in this paper are available at \url{\codedataurl}.

\begin{table}[!t]
\par\vspace{-0.9em}\par
\small
\setlength{\abovecaptionskip}{0.0em}
\caption{\method wins. \method deals with both tasks (T1-T2) for reasoning over TKGs, 
	while representative baselines fail to address both. \method also possesses desirable features (F1-F3) for modeling TKGs.
	TD: TA-DistMult~\cite{DBLP:conf/emnlp/Garcia-DuranDN18}. EG: EvolveGCN~\cite{DBLP:conf/aaai/ParejaDCMSKKSL20}. KE: Know-Evolve~\cite{DBLP:conf/icml/TrivediDWS17}. RN: RE-Net~\cite{DBLP:conf/emnlp/JinQJR20}.}
\centering
\makebox[0.4\textwidth][c]{
\resizebox{0.5\textwidth}{!}{\setlength{\tabcolsep}{0.6mm}
\begin{tabular}{l c c c c c}
& \textbf{TD} & \textbf{EG} & \textbf{KE} & \textbf{RN} & \textbf{\method} \\
	\midrule
\belowrulesepcolor{Highlight}
	\rowcolor{Highlight}
	T1. Modeling evolving network structure			& \cm	&	\cm &		&	\cm	& \boldcheckmark \\ \aboverulesepcolor{Highlight} \midrule[0.2pt]
	\belowrulesepcolor{Highlight}
	\rowcolor{Highlight}
	T2. Modeling event time $ t $					&		&		&	\cm	&		& \boldcheckmark \\
	\rowcolor{Highlight}		
	\enspace • T2-1. Closed-form likelihood\,\&\,expectation	&		&		&	\cm	&		& \boldcheckmark \\
	\rowcolor{Highlight}
	\enspace • T2-2. Flexible approximation of $ p(t) $		&		&		&		&		& \boldcheckmark \\ \aboverulesepcolor{Highlight} \midrule 
	\midrule
	\belowrulesepcolor{LightGray}
	\rowcolor{LightGray}
	F1. Relation-awareness 					& \cm	&		&	\cm	&	\cm	& \boldcheckmark \\
	\rowcolor{LightGray}
	F2. Neighborhood aggregation				& 		&	\cm	&		&	\cm	& \boldcheckmark \\
	\rowcolor{LightGray}
	F3. Recurrent event modeling				& \cm	&	\cm	&	\cm	&	\cm	& \boldcheckmark \\ \aboverulesepcolor{LightGray}
	\bottomrule
\end{tabular}
}
}
\label{tab:salesmatrix}
\vspace{-1.6em}
\end{table}

%% file: 020problem.tex
\textbf{Notations.} A temporal knowledge graph (TKG) $ \g $ is a multi-relational, directed graph with timestamped edges.
We denote a timestamped edge in TKG by a quadruple $ (s,r,o,t) $;
it represents an event between subject entity $ s $ and object entity $ o $, occurring at time $ t $, where
edge type (also called relation) $ r $ denotes the corresponding event type.
In a TKG, we assume no duplicate edges, but there can be multiple edges of the same type between two entities, if they have different timestamps.
For example, a TKG may have both (`u1', `emailed', `u2' `10 am') and (`u1', `emailed', `u2' `12 am').

Let $ (s_n,r_n,o_n,t_n) $ denote an $ n $-th edge among a set of ordered edges.
Given a TKG $ \g $ with $ N $~edges sorted in non-decreasing order of time, we denote it by 
$ \g = \{ (s_n,r_n,o_n,t_n) \}_{n=1}^{N} $ where $ 0 \le t_1 \le t_2 \le \ldots \le t_N $.
We use $ \g_t $ to denote a TKG consisting of events observed at time $ t $, and $ \g_{<t} $ to refer to a TKG with all events observed before time $ t $. We use $ e $ to refer to the event triple $ (s, r, o) $.
We denote vectors by boldface lowercase letters (e.g., $ \vect{c} $), and matrices by boldface capitals (e.g., $ \mat{W} $).

\textbf{Problem: Modeling a TKG.}
Given a TKG $ \g $ with a sequence of observed events $ \{ (s_n,r_n,o_n,t_n) \}_{n=1}^{N} $,
our goal is to model the probability distribution $ p(\g) $.
We assume that events at time $ t $ depend on events that occurred prior to time $ t $, and 
events that happen at the same time are independent of each other, given preceding events.
Based on these assumptions, the joint distribution of TKG $ \g $ can be written as:
\begin{align}\label{eq:kg_prob}
	p(\g) &= \prod_{t}^{} p(\g_t|\priorG) = \prod_{t}^{} \prod_{(s,r,o,t) \in \g_t} p(s,r,o,t |\priorG).
\end{align}
We further decompose the joint conditional probability $ p(s,r,o,t |\priorG) $ in \Cref{eq:kg_prob} as follows. \begin{align}\label{eq:quad_prob}
	p(s,r,o,t |\priorG) &= p(t|s,r,o,\priorG) \cdot p(s,r,o|\priorG)
\end{align}
Note that by modeling the two terms in~\Cref{eq:quad_prob}, we model 
the event time $ p(t|s,r,o,\priorG) $ and the evolving network structure $ p(s,r,o|\priorG) $.
Based on this decomposition, we propose to model a TKG by estimating these two probability terms.

\begin{figure}[t!]
\par\vspace{-0.9em}\par
\centering
\makebox[0.4\textwidth][c]{
\includegraphics[width=0.48\textwidth]{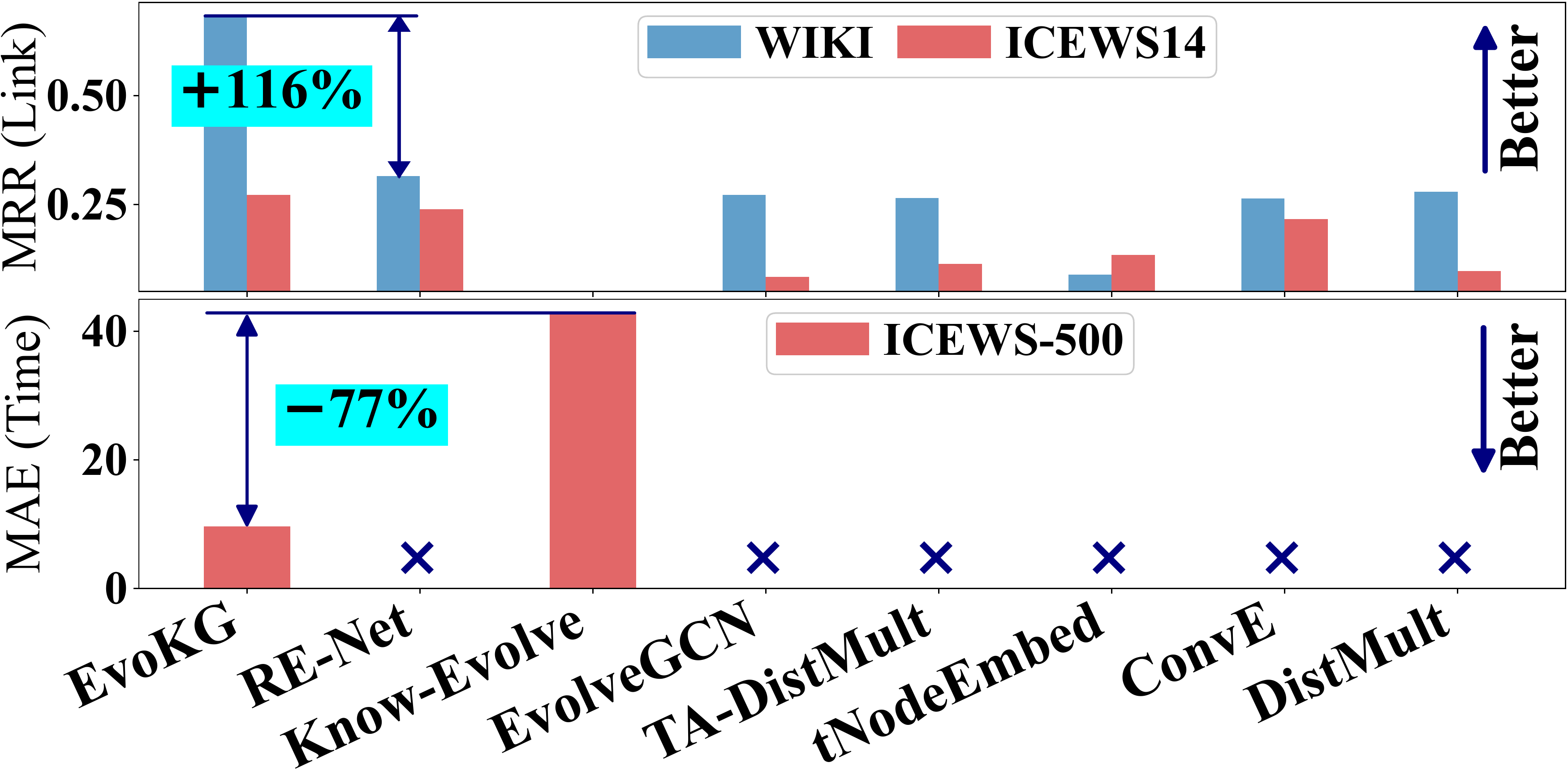}
}
\setlength{\abovecaptionskip}{0.2em}
\captionsetup{width=1.00\linewidth}
\captionof{figure}{\method wins. \method achieves the best link prediction (top) and time prediction (bottom) results. 
	$ \bm{\times} $ indicates that the corresponding method cannot predict event time.}
\label{fig:crownjewel}
\vspace{-1.6em}
\end{figure}

Surprisingly, existing methods for TKGs have focused on modeling either of the two terms, 
but not both at the same time, as summarized in \Cref{tab:salesmatrix}.
Methods that solve only one of the tasks fail to utilize rich information 
that can be learned by addressing the other task:
e.g., methods that do not model the event time 
(e.g., those marked with $ \times $ in \Cref{fig:crownjewel}) cannot predict when events will occur, and
those that only model the event time cannot take the likelihood of an event triple $ (s,r,o) $ into account
when estimating the likelihood of a timestamped event. By unifying these two modeling tasks, we can enable a more accurate reasoning over TKGs.

%% file: 030method.tex
We describe how \method models a TKG by addressing the two problems---modeling event time and evolving network structure.
The symbols used in this paper are listed in \Cref{tab:symbols}.

\subsection{Modeling Event Time}\label{sec:method:temporal}
The temporal patterns of events occurring between various types of entities in a TKG depend on the context of their past interactions.
To capture intricate temporal dependencies present in real-world TKGs, 
we treat the event time $ t $ as a random variable, and model the occurrence of triple $ (s, r, o) $ at time $ t $ 
using temporal point processes (TPPs), which are the dominant paradigm for modeling events that occur at irregular intervals.
Given increasing event times $ \{\ldots,t_{n-1},t_{n},\dots\} $, 
representations in terms of time $ t_{n} $ and the corresponding inter-event time $ \tau_{n} = t_{n} - t_{n-1} $ are isomorphic, and 
we use them interchangeably.

\textbf{Conditional Density Estimation of Event Time.} 
To model the event time, we estimate the conditional probability density $ p^*_e(t)\!=\!p(t|s,r,o,\priorG) $ of event time $ t $,
given an event of type~$ r $ between entities $ s $ and $ o $, and the history $ \priorG $ of all past interactions.
Note that the star symbol $ * $ as in $ p^*_e(t) $ in this paper denotes the dependency on the history $ \priorG $.

More concretely, in order to define $ p^*_e(t) $,
we consider the conditional density of two types of inter-event times $ \tau_{\text{eo}} $ and $ \tau_{\text{min}} $.
Let $ p^{*,e}_{\text{\textsc{eo}}}(t) = p(\tau_{\text{eo}}|s,r,o,\priorG) $ be the conditional density of 
$ \tau_{\text{eo}} $, which is the time that has elapsed since entities $ s $ and $ o $ interacted with each other in their latest event.
Also, let $ p^{*,e}_{\text{\textsc{min}}}(t) = p(\tau_{\text{min}}|s,r,o,\priorG) $ be the conditional density of 
$ \tau_{\text{min}} $, which is defined to be $ \min(\tau_{(s)}, \tau_{(o)}) $, 
where $ \tau_{(s)} $ and $ \tau_{(o)} $ refer to the time 
that has elapsed since $ s $ and $ o $ interacted with any other entity in their latest event.
In other words, $ \tau_{\text{eo}} $ considers how recent the two entities' interaction was, 
while $ \tau_{\text{min}} $ considers when the most recent event happened in either entity's history.
In experiments where we set $ p^*_e(t) $ to be either of these two probabilities, 
we find $ p^{*,e}_{\text{\textsc{min}}}(t) $  and $ p^{*,e}_{\text{\textsc{eo}}}(t) $
to be most effective for predicting event time (\Cref{sec:exp:time}) and temporal links (\Cref{sec:exp:link}), respectively.
Note that $ p^*_e(t) $ can be more generally defined in terms of both conditional densities
to be $ p^*_e(t) = \alpha \cdot p^{*,e}_{\text{\textsc{eo}}}(t) + (1 - \alpha) \cdot p^{*,e}_{\text{\textsc{min}}}(t) $,
where $ \alpha \, (0 \le \alpha \le 1) $ weights each~term, 
and it can also be extended with further conditional densities to model different types of inter-event times.

Importantly, our choice to model event time directly via conditional density estimation differs from existing TPP-based approaches for modeling TKGs~\cite{DBLP:conf/icml/TrivediDWS17,DBLP:conf/akbc/HanM0GT20},
where event times are modeled using the conditional intensity function $ \lambda^*_e(t)\!=\!\lambda(t|s,r,o,\priorG) $,
which represents the rate of events happening, given the history.
In these intensity-based approaches, computing $ p^*_e(t) $ requires integrating $ \lambda^*_e(t) $, and thus 
a major challenge lies in selecting a good parametric form for~$ \lambda^*_e(t) $.
Simple intensity functions (e.g., constant and exponential intensity)
have a closed-form log-likelihood, but they usually have limited expressiveness (e.g., they have a unimodal distribution);
even if they use RNNs to capture rich temporal information, the resulting distribution $ p^*_e(t) $ still has limited flexibility.
More sophisticated ones using neural networks can better capture complex distributions, 
but their log-likelihood and expectation cannot be obtained in closed form, requiring Monte Carlo approximation.
Mixture distributions, on the other hand, are an expressive model for conditional density estimation, 
with the potential to approximate any density, and with closed-form likelihood and expectation. 

Specifically, we use a mixture of log-normal distributions since inter-event times are positive.
Log-normal mixture distributions are defined in terms of mixture weights $ \bm{w} $, means $ \bm{\mu} $, and standard deviations $ \bm{\sigma} $.
An important consideration in employing a log-normal mixture is that the timing of an event in a TKG is affected by 
what has happened before (i.e., $ \priorG $) and what comprises the event triple $ e\!=\!(s,r,o) $.
In light of this, we obtain the three groups of mixture parameters $ \bm{w}_e^* $, $ \bm{\mu}_e^* $, and $ \bm{\sigma}_e^* \in \mathbb{R}^K $,
where the symbols $ e $~and~$ * $ signify these parameters' dependency on the event triple $ e $ and the history $ \priorG $, and
$ K $ denotes the number of mixture components.

To obtain mixture parameters, we learn entity and relation embeddings 
such that they reflect their temporal status (which we describe in the next paragraph), 
as they are influenced by events that occurred over time.
Let $ \emb{t}^{*}_s $, $ \emb{t}^*_o $, and $ \emb{t}^*_r $ denote such temporal embeddings of 
subject $ s $, object $ o $, and relation $ r $, respectively, after processing events prior to time $ t $.
We model the conditional dependence of $ p(\tau|s,r,o,\priorG) $ on $ e\!=\!(s,r,o) $ and $ \priorG $
by concatenating the embeddings of $ s $, $ r $, and $ o $ into 
a context vector $ \vect{c}_e^{*} = [ \emb{t}^*_s \| \emb{t}^*_r \| \emb{t}^*_o ] $, and 
transforming it into the parameters of the log-normal mixture representing $ p(\tau|s,r,o,\priorG) $
using a multilayer perceptron ($ \mathrm{MLP} $) as follows:
\begin{align}\label{eq:lognormal:params}
\bm{w}_e^* = \mathrm{softmax}(\text{\mlp}(\vect{c}_e^{*})),
\bm{\mu}_e^* = \text{\mlp}(\vect{c}_e^{*}),
\bm{\sigma}_e^* = \exp( \text{\mlp}(\vect{c}_e^{*}) )
\end{align}
where $ \mathrm{softmax} $ ensures that mixture weights sum to 1, and $ \exp $ makes standard deviations positive.
With these parameters, \method defines $ p(\tau|s,r,o,\priorG) $ to be
\begin{align}\label{eq:lognormal}
\begin{split}
p(\tau|s,r,o,\priorG) &= p(\tau|\bm{w}_e^*,\bm{\mu}_e^*,\bm{\sigma}_e^*)\\
&=\sum_{k=1}^{K} \frac{(w_e^*)_k}{\tau (\sigma_e^*)_k \sqrt{2\pi}} \exp\left( - \frac{(\log \tau - (\mu_e^*)_k)^2 }{2 {(\sigma_e^*)_k}^2} \right),
\end{split}
\end{align}
which is a valid probability density function as it is nonnegative and integrates to one for $ \tau \in \mathbb{R}_{+} $.

\begin{table}[!t]
\par\vspace{-0.7em}\par
\small
\setlength{\abovecaptionskip}{0.5em}
\caption{Table of symbols.}
\centering
\makebox[0.4\textwidth][c]{
\setlength{\tabcolsep}{0.5mm}
\begin{tabular}{ c p{7.8cm} }
	\toprule
	\textbf{Symbol} & \multicolumn{1}{c}{\textbf{Definition}} \\
	\midrule
	$ (s,r,o,t) $ & directed edge from subject $ s $ to object $ o $, with edge type (relation) $ r $ and timestamp $ t $ \\
	$ e $ & event triple $ (s, r, o) $ \\$ \tau $ & inter-event time (i.e., $ \tau_{n} = t_{n} - t_{n-1} $) \\
	$ \tau_{\text{eo}} $ & elapsed time since entities $ s $ and $ o $ last interacted with each other \\
	$ \tau_{\text{min}} $ & elapsed time since entities $ s $ and $ o $ last interacted with any other entity \\
$ * $ & symbol that signifies that an associated symbol (e.g., $ p^*(\tau) $ and $ \emb{t}^*_i $) depends on the past events \\
	$ p^*_e(\tau) $ & conditional probability density function $ p(\tau|s,r,o,\priorG) $ \\
$ \bm{w}, \bm{\mu}, \bm{\sigma} $ & weights, means, and standard deviations of a log-normal mixture \\
	{$ \emb{t}_i, \emb{s}_i, \emb{t}^*_i, \emb{s}^*_i $} & temporal (structural) embeddings of entity $ i $, with $ * $ reflecting its state after processing events until time $ t $ \\
	{$ \emb{t}_r, \emb{s}_r, \emb{t}^*_r, \emb{s}^*_r $} & temporal (structural) embeddings of relation $ r $, with $ * $ reflecting its state after processing events until time $ t $ \\
$ \emb{t}_{i}^{(\ell,t)}, \emb{s}_{i}^{(\ell,t)} $ & temporal (structural) embeddings of entity $ i $ learned by $ \ell $-th GNN layer at time $ t $\\
	$ \emb{t}_{i}^{(*,t)}, \emb{s}_{i}^{(*,t)} $ & temporal (structural) embeddings of entity $ i $ updated after events until time $ t $ are processed\\
\bottomrule
\end{tabular}
}
\label{tab:symbols}
\vspace{-1.5em}
\end{table}

\textbf{Time-Evolving Temporal Representations.}
Informative context for estimating inter-event time can be constructed by 
summarizing different types of interactions each entity had with others 
into temporal entity embeddings.
Further, how much time elapsed since the latest event gives useful information for learning such temporal representations.
To this end, we utilize the neighborhood aggregation framework of relation-aware graph neural networks (GNNs).
Specifically, we extend R-GCN~\cite{DBLP:conf/esws/SchlichtkrullKB18} 
such that the aggregation can take inter-event time $ \tau_{i,j} $ between entities $ i $ and $ j $ into account.
Given concurrent events $ \g_t $, we summarize entity $ i $'s interaction with others in $ \g_t $ as follows:
\begin{align}\label{eq:rgcn:temporal}
	\emb{t}_{i}^{(\ell+1,t)} = \sigma \bigg( \sum_{r \in \mathcal{R}} \sum_{j \in \mathcal{N}_t^{(i,r)}} \frac{1}{\nu_{i,j}} \cdot \mat{W}_r^{\ell}  \emb{t}_{j}^{(\ell,t)} + \mat{W}_0^{\ell} \emb{t}_{i}^{(\ell,t)} \bigg)
\end{align}
where $ \emb{t}_{i}^{(\ell,t)} $ denotes the temporal embeddings of entity $ i $ 
learned by $ \ell $-th layer of the extended R-GCN by aggregating events in $ \g_t $;
$ \nu_{i,j} $ is a factor to consider the inter-event time, 
which we define to be $ \nu_{i,j} = \log \tau_{i,j} $;
$ \mathcal{R} $ is the set of relations; 
$ \mathcal{N}_t^{(i,r)} $ is entity $ i $'s concurrent neighbors at time $ t $, connected via an edge of type $ r $;
$ \mat{W}_r^{\ell} $ and $ \mat{W}_0^{\ell} $ are the weight matrices in the $ \ell $-th layer for relation $ r $ and self-loop, respectively.
Then with $ L $ layers in total, $ \emb{t}_i^{L} $ summarizes entity~$ i $'s temporal interactions in the $ L $-hop neighborhood.
The initial temporal embeddings $ \emb{t}_i^{(0,t)} $ are set to 
the static representation~$ \emb{t}_i $ that \method learns to capture the temporal characteristics of entities, 
i.e., $ \emb{t}_i^{(0,t)} = \emb{t}_i $ for any time $ t $.

To model the dynamics of temporal updates,
the context for modeling inter-event time should reflect the changes made by new events.
Given $ \emb{t}_i^{(L,t)} $ which summarizes the temporal interaction patterns from concurrent events at time $ t $,
\method learns time-evolving dynamics from the evolution of $ \emb{t}_i^{(L,t)} $ over time, 
by using recurrent neural networks {\rnn}\textsubscript{te} for temporal entity representation learning:
\begin{align}\label{eq:rnn:temporal:entity}
\emb{t}_i^{(*,t)} = \rnn_{\text{te}} \big(\emb{t}_i^{(L,t)}, \emb{t}_i^{(*,t-1)} \big)
\end{align}
where $ \emb{t}_i^{(L,t)} $ is the input to {\rnn}s at each time; $ \emb{t}_i^{(*,0)} $ is zero-initialized; and 
$ \emb{t}_i^{(*,t)} $ is the temporal embedding of entity~$ i $ updated after events until time $ t $ are processed.
In this framework, as aggregating incoming and outgoing neighbors captures sending and receiving patterns between entities,
\method aggregates neighborhood in both directions
to learn embeddings that reflect different interaction patterns,
which are then processed by {\rnn}\textsubscript{te} to be used in~the~context~$ \vect{c}_e^{*} $.

Next, \method considers the concurrent events $ \g_t^r $ that have relation~$ r $, and 
takes the average of the temporal embeddings of the entities in $ \g_t^r $ 
to provide it as the context $ \emb{t}_r^{t} $ to {\rnn}\textsubscript{tr},
which learns the temporal embedding $ \emb{t}_r^{(*,t)} $ of relation $ r $ at time $ t $:
\begin{align}\label{eq:rnn:temporal:rel}
	\emb{t}_r^{(*,t)} = \rnn_{\text{tr}} \big(\emb{t}_r^{t}, \emb{t}_r^{(*,t-1)} \big).
\end{align}
For brevity, we use the notation $ \emb{t}_i^* = \emb{t}^{(*,t)}_i $ and $ \emb{t}_r^{*} = \emb{t}_r^{(*,t)} $.

\vspace{-0.9em}
\subsection{Modeling Evolving Network Structure}\label{sec:method:structural}

As new events occur, TKGs evolve structurally and the dynamics between entities also change over time.
For instance, companies that did not work together may start to collaborate at some point to work on the same project, 
and this change may influence the communication patterns between them and related entities in the TKG.
We capture this intricate structural dynamics by modeling the conditional probability $ p(s,r,o|\priorG) $ of an event triple $ (s,r,o) $.

\textbf{Conditional Density Estimation of Event Triple.} 
To model $ p(s,r,o|\priorG) $, we learn the embeddings of entities and relations (which we discuss in the next paragraph), 
which capture their time-evolving structural dynamics.
{For flexibility, we learn these embeddings separately from temporal embeddings discussed in \Cref{{sec:method:temporal}}.}
Let $ \emb{s}_i $ and $ \emb{s}_r $ denote the static structural embeddings of entity~$ i $ and relation $ r $, and 
let $ \emb{s}^{*}_i $ and $ \emb{s}^{*}_r $ be the structural embeddings of entity~$ i $ and relation $ r $ obtained by processing events until time $ t $.
{We concatenate static and dynamic embeddings and denote them using $ \overbar{\emb{s}}^{*}_i = [ \emb{s}^{*}_i \| \emb{s}_i ] $ and $ \overbar{\emb{s}}^{*}_r = [ \emb{s}^{*}_r \| \emb{s}_r ] $.}
Then \method summarizes the past events $ \priorG $, which $ p(s,r,o|\priorG) $ is conditioned on,
by the graph-level representation $ \overbar{\emb{g}}^{*} $, which \method obtains 
via an element-wise max pooling over the structural embeddings of all entities,~i.e., \begin{align}
\overbar{\emb{g}}^{*}\!=\!\max(\{ \overbar{\emb{s}}^{*}_i \mid i \in \mathrm{entities(}\priorG\text{)} \} ).
\end{align}
Based on these representations, we decompose $ p(s,r,o|\priorG) $ to be
\begin{align}\label{eq:triple_prob}
p(s,r,o|\priorG) = p(o|s,r,\priorG) \cdot p(r|s,\priorG) \cdot p(s|\priorG)
\end{align}
and parameterize each term separately, as follows:
\begin{align}\label{eq:triple_prob:parameterization}
p(o|s,r,\priorG) &= \mathrm{softmax}\left( \text{\mlp} \left( [ \overbar{\emb{s}}^{*}_s \| \overbar{\emb{s}}^{*}_r \| \overbar{\emb{g}}^{*} ] \right) \right),\\
p(r|s,\priorG) &= \mathrm{softmax}\left( \text{\mlp} \left( [ \overbar{\emb{s}}^{*}_s \| \overbar{\emb{g}}^{*} ] \right) \right),\\
p(s|\priorG) &= \mathrm{softmax}\left( \text{\mlp} \left( [ \overbar{\emb{g}}^{*} ] \right) \right).
\end{align}

\textbf{Time-Evolving Structural Representations.}
An effective modeling of $ p(s,r,o|\priorG) $ based on the above parameterization
depends~on learning informative context that reflects how structural dynamics between entities have changed over time.
As with learning temporal embeddings, neighborhood aggregation of GNNs and recurrent event modeling using {\rnn}s 
provide an effective framework to capture this complex structural evolution. 
Thus, we adapt the framework used for event time modeling in \Cref{sec:method:temporal} for learning time-evolving structural embeddings.

Let $ \emb{s}_{i}^{(\ell,t)} $ denote the structural embeddings of entity~$ i $ 
learned by $ \ell $-th R-GCN layer by aggregating concurrent events $ \g_t $.
As before, we set the initial structural embeddings $ \emb{s}_{i}^{(0,t)} $ to $ \emb{s}_i $ for each time~$ t $.
Given embeddings $ \emb{s}_{i}^{(\ell,t)} $ for all entities, $ \emb{s}_{i}^{(\ell+1,t)} $ is learned using \Cref{eq:rgcn:temporal},
where $ \emb{t}_{i}^{(\ell,t)} $ is replaced by $ \emb{s}_{i}^{(\ell,t)} $, and 
$ \nu_{i,j} $ is set to the neighborhood size $ |\mathcal{N}_t^{(i,r)}| $.
The structural relation embedding $ \emb{s}_r^{t} $ at time $ t $ is constructed using concurrent events in the same way as in the temporal case.
Then \method learns the time-evolving structural embeddings $ \emb{s}_i^{(*,t)} $ and $ \emb{s}_r^{(*,t)} $
using {\rnn}\textsubscript{se} and {\rnn}\textsubscript{sr} as follows.
\begin{align}\label{eq:rnn:structural}
\emb{s}_i^{(*,t)} = \rnn_{\text{se}} \big(\emb{s}_i^{(L,t)}, \emb{s}_i^{(*,t-1)} \big),\,\,
\emb{s}_r^{(*,t)} = \rnn_{\text{sr}} \big(\emb{s}_r^{t}, \emb{s}_r^{(*,t-1)} \big)
\end{align}
For brevity, we use the notation $ \emb{s}_i^* = \emb{s}^{(*,t)}_i $ and $ \emb{s}_r^{*} = \emb{s}_r^{(*,t)} $.

\vspace{-0.9em}
\subsection{Parameter Learning}
\vspace{-0.3em}
\indent\indent
\textbf{Loss Function.}
Let $ \mathcal{L}_{\text{iet}} $ and $ \mathcal{L}_{\text{triple}} $ denote 
the negative log-likelihood (NLL) of the inter-event time and an event triple, respectively.
Based on our problem formulation and modeling choices, 
the two NLLs of a quadruple $ q\!=\!(s,r,o,t) $ (i.e., a timestamped event in a TKG) are obtained as follows.
\begin{align}
\mathcal{L}_{\text{iet}}(q) &= - \log p(t|s,r,o,\priorG) = - \log p(\tau|\bm{w}_e^*,\bm{\mu}_e^*,\bm{\sigma}_e^*) \\
\begin{split}
\mathcal{L}_{\text{triple}}(q) &= - \log p(s,r,o|\priorG) \\
&= - \log p(o|s,r,\priorG) - \log p(r|s,\priorG) - \log p(s|\priorG)
\end{split}
\end{align}
We optimize \method by minimizing the loss $ \mathcal{L} $ containing both NLLs for all events in the training set:
\begin{align}\label{eq:loss}
\mathcal{L} &= \sum_{t}^{} \sum_{q=(s,r,o,t) \in \g_t} \lambda_{1}\,\mathcal{L}_{\text{iet}}(q) + \lambda_{2}\,\mathcal{L}_{\text{triple}}(q)
\end{align}
where $ \lambda_{1} $ and $ \lambda_{2} $ control the importance of each loss term.

\textbf{Learning Algorithm.}
Since there exist intricate relational and temporal dependencies among events in TKGs,
it is not optimal to decompose events into independent sequences for an efficient training, as we lose relational information.
At the same time, since a TKG may cover a long period of time, 
keeping track of the entire history for each entity can incur prohibitively high computation and memory cost, 
especially when learning graph-contextualized representations for entities and relations.
To address these challenges, 
we organize events by their timestamps and process concurrent events in parallel,
while truncating backpropagation every $ b $ time steps (Algorithm~\ref{alg:learning}).
As experimental results show, this enables an accurate and efficient parameter learning, 
which outperforms the best baseline in terms of both prediction accuracy and efficiency.

{
\fontsize{9pt}{10}\selectfont
\begin{algorithm}[!t]
\DontPrintSemicolon
\SetNoFillComment
\SetKwComment{Comment}{$\triangleright$\ }{}
\KwIn{TKG $ \g $ with training data, TKG $ \g' $ with validation data, maximum number of epochs $ max\_epochs $, number $ L $ of R-GCN layers, patience $ p $, number of time steps $ b $ for truncated backpropagation.}
$ epoch \gets 1 $\\
\Repeat{epoch $ = $ max\_epochs \normalfont{or no improvement in validation performance for $ p $ consecutive times}}{
\ForEach{$t \in \text{Timestamps}(\g)$}{
		\If{$ t > 0 $}{
Compute the loss $ \mathcal{L}_t $ for concurrent events in $ \g_t $ based on \Cref{eq:loss}\\
			Optimize model parameters and truncate backpropagation every $ b $ time steps
		}
		
		\ForEach(\tcc*[h]{{\normalfont \textit{\!\!{executed in parallel}\!\!}}}){$i\! \in\! \text{Entities}\,(\g_t)$}{
			Compute $ \emb{t}^{(L,t)}_i $ and $ \emb{t}^{*}_i $ using \cref{eq:rgcn:temporal,eq:rnn:temporal:entity}\\
			Compute $ \emb{s}^{(L,t)}_i $ and $ \emb{s}^{*}_i $ using a modified version of \cref{eq:rgcn:temporal} and \cref{eq:rnn:structural}\\
		}
		\ForEach(\tcc*[h]{{\normalfont \textit{\!{executed in parallel}\!}}}){$r \in \text{Rels}\,(\g_t)$}{
			Compute $ \emb{t}^{*}_r $ and $ \emb{s}^{*}_r $ using \cref{eq:rnn:temporal:rel,eq:rnn:structural}\\
		}
	}
	Evaluate the validation performance for events in $ \g' $\\
$ epoch \gets epoch + 1 $
}
\caption{Parameter Learning}
\label{alg:learning}
\end{algorithm}
}

\begin{table}[t!]
\par\vspace{-1.0em}\par
\small
\setlength{\tabcolsep}{1.0mm}
\setlength{\abovecaptionskip}{0.0em}
\captionsetup{width=1.07\linewidth}
\caption{Statistics of real-world TKGs. Time interval denotes the minimum duration between two temporally adjacent events.}
\label{tab:datasets}
\centering
\makebox[0.4\textwidth][c]{
	\begin{tabular}{c r r r r r c}
		\toprule
		\textbf{Dataset} & \textbf{\makecell{\# Train\\Edges}} & \textbf{\makecell{\# Valid\\Edges}} & \textbf{\makecell{\# Test\\Edges}} & \textbf{\# Entities} & \textbf{\makecell{\# Rel-\\ations}} & \textbf{\makecell{Time\\Interval}} \\ \midrule
		\icews & 373,018 & 45,995 & 49,545 & 23,033 & 256 & 24 hours \\ 
		\icewsFourteen & 275,367 & 48,528 & 341,409 & 12,498 & 260 & 24 hours \\ 
		\icewsSmall & 184,725 & 32,292 & 228,648 & 500 & 256 & 24 hours \\ 
		\gdelt & 1,734,399 & 238,765 & 305,241 & 7,691 & 240 & 15 minutes \\ 
		\wiki & 539,286 & 67,538 & 63,110 & 12,554 & 24 & 1 year \\
		\yago & 161,540 & 19,523 & 20,026 & 10,623 & 10 & 1 year \\
		\bottomrule
\end{tabular}}
\vspace{-1.0em}
\end{table}

%% file: 040experiments.tex
In experiments, we answer the following research questions.
\begin{itemize}[nosep,leftmargin=1em]
	\item \textbf{[RQ1]} How accurately can \method estimate the event time?
	\item \textbf{[RQ2]} How accurately can \method predict temporal links?
	\item \textbf{[RQ3]} How efficient is \method in terms of training and inference?
	\item \textbf{[RQ4]} How do different parameter settings and event time modeling affect \method's performance?
\end{itemize}
After describing the datasets (\Cref{sec:exp:data}), we present results for the above research questions (\Cref{sec:exp:time,sec:exp:link,sec:exp:efficiency,sec:exp:ablation}).
Experimental settings are provided in~\Cref{sec:appendix}.

\subsection{Temporal Knowledge Graph Data}\label{sec:exp:data}

We use five real-world TKGs that have been widely used in previous studies:
\icews~\cite{boschee2015icews}, \icewsFourteen~\cite{DBLP:conf/icml/TrivediDWS17}, 
\gdelt~\cite{leetaru2013gdelt}, \wiki~\cite{DBLP:conf/www/LeblayC18}, and \yago~\cite{DBLP:conf/cidr/MahdisoltaniBS15}.
ICEWS (Integrated Crisis Early Warning System) and 
\gdelt (Global Database of Events, Language, and Tone) are event-based TKGs;
\wiki and \yago are knowledge bases with temporally associated facts.
Statistics of these TKGs are presented in \Cref{tab:datasets}.
We order these datasets by timestamps, and split each one into training, validation, and test sets, as shown in~\Cref{tab:datasets}.
We also use \icewsSmall~\cite{DBLP:conf/icml/TrivediDWS17} for experiments on event time prediction, 
which is a TKG constructed from ICEWS data, containing a smaller number of nodes than \icews,
since some previous studies reported results only on \icewsSmall without releasing code.

\begin{figure}[t!]
\centering
\makebox[0.4\textwidth][c]{
\includegraphics[width=0.525\textwidth]{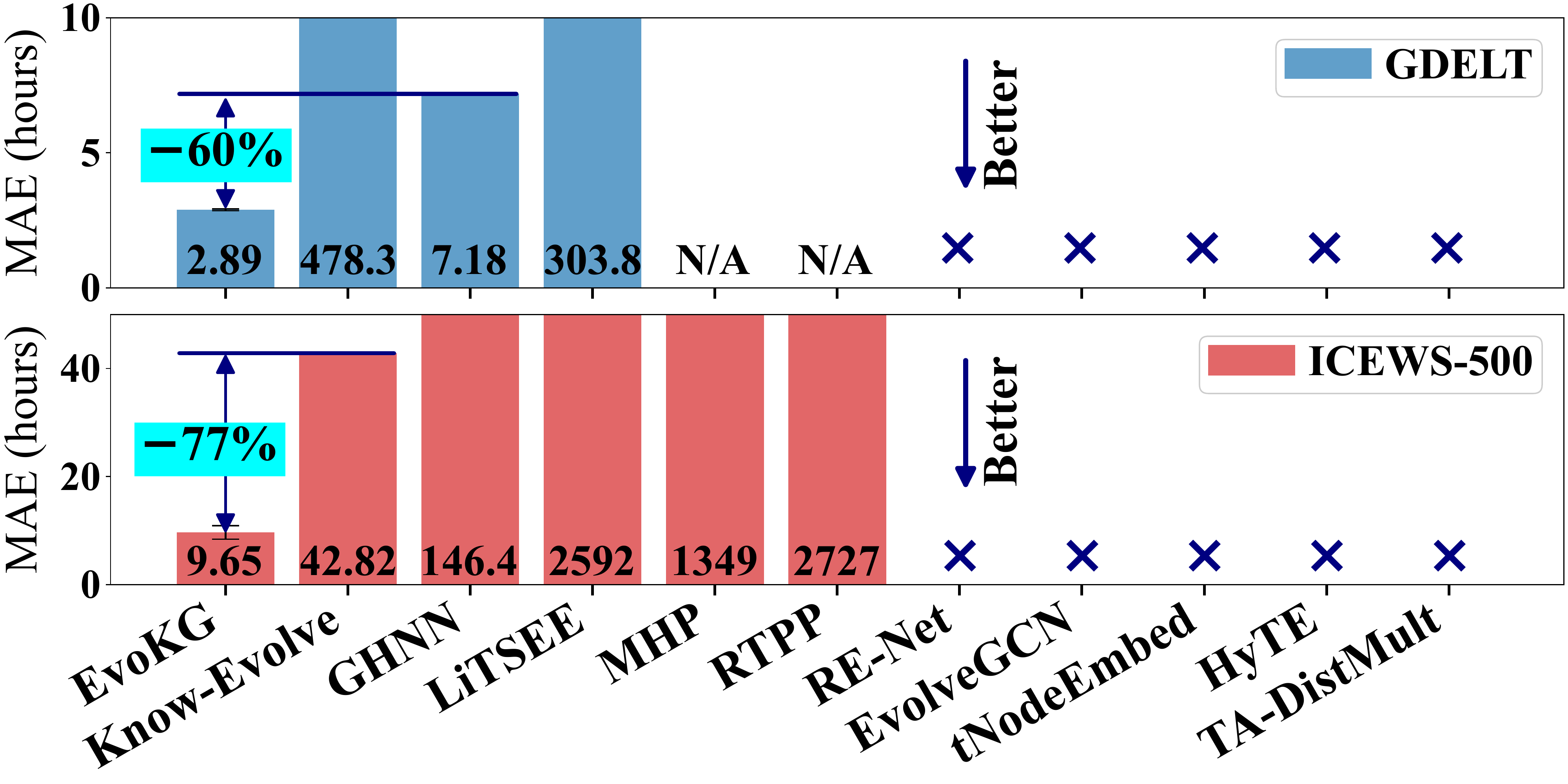}
}
\setlength{\abovecaptionskip}{0.2em}
\captionsetup{width=1.0\linewidth}
\captionof{figure}{\method achieves the best event time prediction results, 
with up to 77\% less MAE than the second best method; all improvements are statistically significant with $ p $-value < 0.05.
Note that many methods for TKGs (marked by $ \bm{\times} $) cannot predict event time.
N/A denotes results are unavailable.
}
\label{fig:exp:timepred}
\end{figure}

\subsection{Event Time Prediction (RQ1)}\label{sec:exp:time}

\indent\indent
\textbf{Task Description.}
Given an event triple $ e\!=\!(s, r, o) $ and the history $ \priorG $, the goal is to predict when the event $ e $ will happen.
Specifically, the time of an event triple $ e $ is estimated to be 
the expected value of the time that event $ e $ occurs, given the history.
Thanks to the use of a mixture distribution,
in \method, this expectation is obtained in a closed form by
\begin{align}
\mathbb{E}_{\tau \sim p^*_e(\tau)}(\tau) = \sum\nolimits_{k} (w_e^*)_k \exp\left( (\mu_e^*)_k + \nicefrac{((s_e^*)_k)^2}{2} \right).
\end{align}
On the other hand, other approaches, such as GHNN~\cite{DBLP:conf/akbc/HanM0GT20}, 
need to approximate the integral to compute the expected value by using Monte Carlo,
as~they do not have a close-form solution.
We report MAE (mean absolute error), which is the average of the absolute difference between the predicted and true time in hours.
Lower MAE indicates higher prediction accuracy.

\textbf{Baselines.} 
We compare \method against three existing methods for modeling TKGs with the ability to predict event time:
\knowevolve~\cite{DBLP:conf/icml/TrivediDWS17}, \ghnn~\cite{DBLP:conf/akbc/HanM0GT20}, and \litsee~\cite{DBLP:journals/corr/abs-1911-07893}.
\knowevolve and \ghnn model event time based on temporal point process (TPP) framework.
While there exist several other methods for modeling TKGs, 
they are unable to forecast event time, thus they cannot be used for this evaluation.
We also report the result of two other baselines used in \cite{DBLP:conf/icml/TrivediDWS17}, 
\mhp (Multi-dimensional Hawkes Process) and \rtpp (Recurrent Temporal Point Process).
\mhp models dyadic entity interactions as multi-dimensional Hawkes process: 
an entity pair constitutes an event, and \mhp learns when each event occurs, without taking relations (event types) into account.
\rtpp is a simplified version of RMTPP~\cite{DBLP:conf/kdd/DuDTUGS16}, 
which estimates the conditional intensity function of an event by using a global RNN. Relations are also considered in \rtpp.

\textbf{Results.}
\Cref{fig:exp:timepred} reports the event time prediction accuracy on \icewsSmall and \gdelt\label{key}.
Results of \knowevolve, \rtpp, and \mhp are obtained from~\cite{DBLP:conf/icml/TrivediDWS17} 
(except that we obtained \knowevolve's result on \gdelt using the reference implementation), and 
those of \litsee and \ghnn are taken from~\cite{DBLP:conf/akbc/HanM0GT20}.
Notably, most TKG methods in \Cref{fig:exp:timepred} are marked with $ \times $ due to their inability to estimate event time.
Also, the results of \rtpp and \mhp are not available (marked with `\NA')
as their implementation is not publicly available.
In \method, we used $ \tau_{\text{min}} $ to model the inter-event time.
{In experiments, methods are updated using the observed graph snapshot at each time step to make future predictions.}

Results show that \method consistently outperforms all existing approaches, with up to 77\% less MAE than the second-best method.
We conduct one-sample t-tests and verify that all improvements over baselines are statistically significant with $ p $-value < 0.05.
Graph-based methods (\method, \knowevolve, and \ghnn), 
which learn the temporal patterns of events by utilizing information from the neighborhood,
perform much better than simpler baselines (\rtpp and \mhp), which model event time based only on direct interactions between entities.
Also, TPP-based \knowevolve and \ghnn outperform \litsee,
a non-TPP approach which incorporates time information by adding a temporal component into entity embeddings.
\method achieves the best event time prediction results
by modeling event time using mixture distributions, which are much more flexible and expressive than those used by existing methods.

\vspace{-1.4em}
\subsection{Temporal Link Prediction (RQ2)}\label{sec:exp:link}

\indent\indent
\textbf{Task Description.} Given a test quadruple $ q\!=\!(s, r, o, t) $ and the history $ \priorG $, 
we create a perturbed quadruple $ q'\!=\!(s, r, o', t) $ by replacing $ o $ with every other entity $ o' $ in the graph, and compute the score of $ q' $.
We then sort all perturbed quadruples in descending order of the score and report the rank of the ground truth quadruple~$ q $.
We report MRR (mean reciprocal rank), which is the average of the reciprocal of the ground truth $ q $'s rank,
and Hits@\{3,10\}, which is the percentage of correct entities in the top 3 and 10 predictions.
For both metrics, higher values indicate better link prediction results.

\textbf{Baselines.}
We compare \method against the following baselines for both static and temporal KG reasoning.
(1) DistMult~\cite{DBLP:journals/corr/YangYHGD14a}, R-GCN~\cite{DBLP:conf/esws/SchlichtkrullKB18}, 
ConvE~\cite{DBLP:conf/aaai/DettmersMS018}, and RotatE~\cite{DBLP:conf/iclr/SunDNT19} are methods for static KG reasoning.
They are applied to a static, cumulative graph constructed from events in the training data, where edge timestamps are ignored.
(2) TA-DistMult~\cite{DBLP:conf/emnlp/Garcia-DuranDN18} and HyTE~\cite{DBLP:conf/emnlp/DasguptaRT18} are methods 
for temporal KG reasoning in an interpolation setting.
(3) dyngraph2vecAE~\cite{DBLP:journals/kbs/GoyalCC20}, tNodeEmbed~\cite{DBLP:conf/ijcai/SingerGR19}, 
EvolveGCN~\cite{DBLP:conf/aaai/ParejaDCMSKKSL20}, and GCRN~\cite{DBLP:conf/iconip/SeoDVB18} are methods 
for reasoning over homogeneous graphs in an extrapolation setting.
(4) Know-Evolve~\cite{DBLP:conf/icml/TrivediDWS17}, DyRep~\cite{DBLP:conf/iclr/TrivediFBZ19}, and 
RE-Net~\cite{DBLP:conf/emnlp/JinQJR20} are methods for temporal KG reasoning in an extrapolation setting.
\ghnn~\cite{DBLP:conf/akbc/HanM0GT20} is not included as the implementation is not available.
Also, results reported in~\cite{DBLP:conf/akbc/HanM0GT20} were obtained 
after applying its own filtering criteria, where \ghnn achieved similar results to \renet.

\textbf{Results.}
\Cref{tab:exp:linkpred} provides link prediction results on five TKGs.
Results of baselines are obtained from~\cite{DBLP:conf/emnlp/JinQJR20}.
In \method, we used $ \tau_{\text{eo}} $ to model the inter-event time.
{In experiments, after making predictions at each time step, methods are updated using the observed graph snapshot.}
\method outperforms all existing approaches across different datasets, achieving up to 116\% higher MRR than the best baseline,
except on \gdelt, where \method achieves similar performance to the best baseline.
It is noteworthy that an improvement over baselines is the most significant on \wiki and \yago, 
which contains much more events occurring at relatively regular intervals.
By modeling event time, \method can predict such temporal patterns accurately.
Among baselines, static methods in the first four rows perform worse than the best temporal baseline, \renet, 
as they do not consider temporal factors.
At the same time, some temporal methods, such as dyngraph2vecAE and EvolveGCN, often perform worse than static methods, 
even though they are designed to take temporal evolution of dynamic networks into account.
This indicates that incorporating temporal factors needs to be done carefully to avoid introducing additional noise.
\knowevolve and \dyrep are the two existing methods based on temporal point processes.
While they can be used for temporal link prediction,
they are not effective for predicting links, even after applying an MLP decoder to their embeddings, 
as they focus on modeling just $ p(t|s,r,o,\priorG) $, and
thus do not explicitly learn the evolving network structure by modeling $ p(s,r,o|\priorG) $ as in \method.
By modeling event time and network structure simultaneously, 
\method outperforms various existing methods in predicting temporal links and event times.

\begin{table*}[t!]
\setlength{\tabcolsep}{1.0mm}
\setlength{\abovecaptionskip}{5pt}
\captionsetup{width=1.0\linewidth}
\caption{\method outperforms existing methods in terms of temporal link prediction in most cases, 
achieving up to 116\% higher MRR (mean reciprocal rank) on real-world TKGs. 
Best results are in bold, and second best results are underlined.}
\label{tab:exp:linkpred}
\centering
\makebox[0.4\textwidth][c]{
	\resizebox{1.00\textwidth}{!}{
		\begin{tabular}{ll|rrr|rrr|rrr|rrr|rrr}
			\toprule
			\multirow{2}{*}{} & \multicolumn{1}{l}{\multirow{2}{*}{\vspace{-2mm}\textbf{Method}}} &\multicolumn{3}{c}{\textbf{\icewsFourteen}} &\multicolumn{3}{c}{\textbf{\icews}} &\multicolumn{3}{c}{\textbf{\wiki}} &\multicolumn{3}{c}{\textbf{\yago}} &\multicolumn{3}{c}{\textbf{\gdelt}} \\
			\cmidrule(lr){3-5} \cmidrule(lr){6-8} \cmidrule(lr){9-11} \cmidrule(lr){12-14} \cmidrule(lr){15-17} 
			& &MRR &H@3 &H@10 &MRR &H@3 &H@10 &MRR &H@3 &H@10 &MRR &H@3 &H@10 &MRR &H@3 &H@10 \\
			\midrule
\multirow{4}{*}{\rotatebox{90}{Static}}
			&DistMult &9.72 &10.09 &22.53 &13.86 &15.22 &31.26 &27.96 &32.45 &39.51 &44.05 &49.70 &59.94 &8.61 &8.27 &17.04 \\
			&R-GCN &15.03 &16.12 &31.47 &15.05 &16.49 &29.00 &13.96 &15.75 &22.05 &27.43 &31.24 &44.75 &12.17 &12.37 &20.63 \\
			&ConvE &21.64 &23.16 &38.37 &22.56 &25.41 &41.67 &26.41 &30.36 &39.41 &41.31 &47.10 &59.67 &18.43 &19.57 &32.25 \\
			&RotateE &9.79 &9.37 &22.24 &11.63 &12.31 &28.03 &26.08 &31.63 &38.51 &42.08 &46.77 &59.39 &3.62 &2.26 &8.37 \\
			\cmidrule{1-17}
\multirow{10}{*}{\rotatebox{90}{Temporal}}
			&TA-DistMult &11.29 &11.60 &23.71 &15.62 &17.09 &32.21 &26.44 &31.36 &38.97 &44.98 &50.64 &61.11 &10.34 &10.44 &21.63 \\
			&HyTE &7.72 &7.94 &20.16 &7.41 &7.33 &16.01 &25.40 &29.16 &37.54 &14.42 &39.73 &46.98 &6.69 &7.57 &19.06 \\
			&dyngraph2vecAE &6.95 &8.17 &12.18 &1.36 &1.54 &1.61 &2.67 &2.75 &3.00 &0.81 &0.74 &0.76 &4.53 &1.87 &1.87 \\
			&tNodeEmbed &13.36 &13.13 &24.31 &7.21 &7.64 &15.75 &8.86 &10.11 &16.36 &3.82 &3.88 &8.07 &12.97 &12.61 &21.22 \\
			&EvolveGCN &8.32 &7.64 &18.81 &10.31 &10.52 &23.65 &27.19 &31.35 &38.13 &40.50 &45.78 &55.29 &6.54 &5.64 &15.22 \\
			&Know-Evolve &0.05 &0.00 &0.10 &0.11 &0.00 &0.47 &0.03 &0.00 &0.04 &0.02 &0.00 &0.01 &0.11 &0.02 &0.10 \\
			&Know-Evolve+MLP &16.81 &18.63 &29.20 &7.41 &7.87 &14.76 &10.54 &13.08 &20.21 &5.23 &5.63 &10.23 &15.88 &15.69 &22.28 \\
			&DyRep+MLP &17.54 &19.87 &30.34 &7.82 &7.73 &16.33 &10.41 &12.06 &20.93 &4.98 &5.54 &10.19 &16.25 &16.45 &23.86 \\
			&R-GCRN+MLP &21.39 &23.60 &38.96 &23.46 &26.62 &41.96 &28.68 &31.44 &38.58 &43.71 &48.53 &56.98 &18.63 &19.80 &32.42 \\
&\renet &\secondbest{23.91} &\secondbest{26.63} &\secondbest{42.70} &\secondbest{26.81} &\secondbest{30.58} &\secondbest{45.92} &\secondbest{31.55} &\secondbest{34.45} &\secondbest{42.26} &\secondbest{46.37} &\secondbest{51.95} &\secondbest{61.59} &\textbf{19.44} &\textbf{20.73} &\secondbest{33.81} \\
			\midrule\midrule
			\belowrulesepcolor{Gray}
			\rowcolor{Gray}
			&\textbf{\method}
			&\makecell[r]{\textbf{27.18}\\$\pm$0.001} &\makecell[r]{\textbf{30.84}\\$\pm$0.001}
			&\makecell[r]{\textbf{47.67}\\$\pm$0.001}
			&\makecell[r]{\textbf{29.28}\\$\pm$0.002} &\makecell[r]{\textbf{33.94}\\$\pm$0.004}
			&\makecell[r]{\textbf{50.09}\\$\pm$0.002}
			&\makecell[r]{\textbf{68.03}\\$\pm$0.031} &\makecell[r]{\textbf{79.60}\\$\pm$0.036}
			&\makecell[r]{\textbf{85.91}\\$\pm$0.063}
			&\makecell[r]{\textbf{68.59}\\$\pm$0.003} &\makecell[r]{\textbf{81.13}\\$\pm$0.005}
			&\makecell[r]{\textbf{92.73}\\$\pm$0.009}
			&\makecell[r]{\secondbest{19.28}\\$\pm$0.001} &\makecell[r]{\secondbest{20.55}\\$\pm$0.001}
			&\makecell[r]{\textbf{34.44}\\$\pm$0.002}\\
\aboverulesepcolor{Gray}
			\bottomrule
		\end{tabular}
}}\end{table*}

\begin{figure}[t!]
\centering
\includegraphics[width=0.425\textwidth]{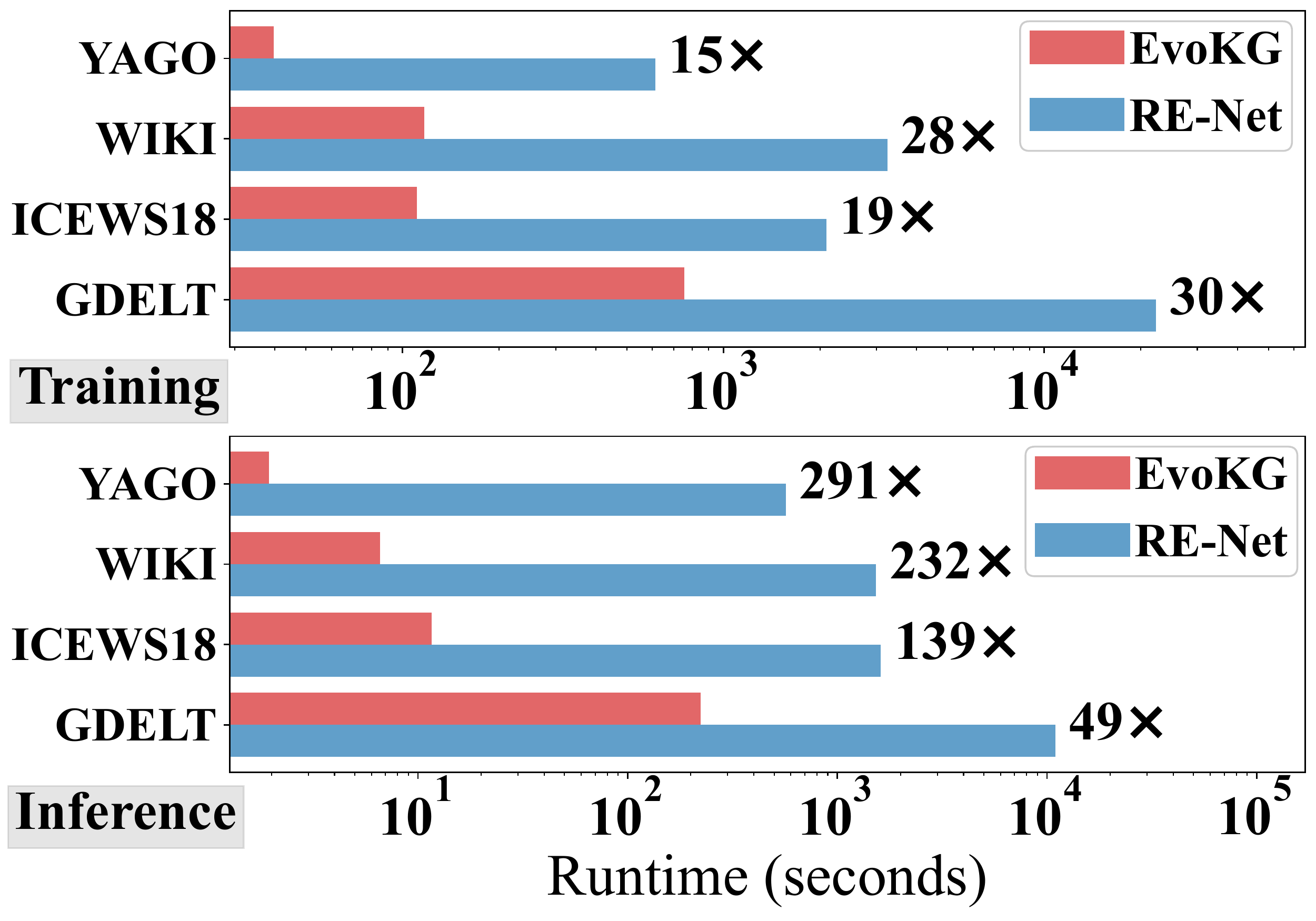}
\setlength{\abovecaptionskip}{0.5em}
\captionsetup{width=1.025\linewidth}
\captionof{figure}{\method performs training (top) and inference (bottom) up to $ 30\times $ and $ 291\times $ faster than \renet.}
\label{fig:exp:efficiency}
\vspace{-1.0em}
\end{figure}

\vspace{-0.9em}
\subsection{Efficiency (RQ3)}\label{sec:exp:efficiency}
\vspace{-0.1em}

\indent\indent
\textbf{Setup.} We compare \method against \renet, the best performing baseline method for temporal link prediction, 
in terms of model training and inference speed.
We evaluate the training speed by measuring the time taken to train one epoch, 
and the inference speed by measuring the time taken to evaluate the entire test data in terms of $ p(s,r,o|\priorG) $.

\textbf{Results.} \Cref{fig:exp:efficiency} shows the time taken for training (top) and inference (bottom) over four TKGs.
The training speed for \method is 23$ \times $ on average, and up to 30$ \times $, faster than \renet.
In making inferences, \method is 177$ \times $ on average, and up to 291$ \times $, faster than \renet.
This is because \renet's design for handling events results in a lot of repeated computations for neighborhood aggregation and processing event history.
The difference in runtime is even more pronounced in making inferences since \renet processes event quadruples individually during inference.
On the other hand, \method processes concurrent events simultaneously, effectively reducing redundant operations.
As a result, \method performs both tasks much more efficiently than \renet.

\vspace{-0.9em}
\subsection{Ablation Study (RQ4)}\label{sec:exp:ablation}

\subsubsection{Parameter Sensitivity}\label{sec:exp:ablation:parameter}

\begin{figure}[t!]
\centering
\makebox[0.4\textwidth][c]{
	\begin{subfigure}[b]{0.55\textwidth}
		\centering
		\includegraphics[width=1.0\textwidth]{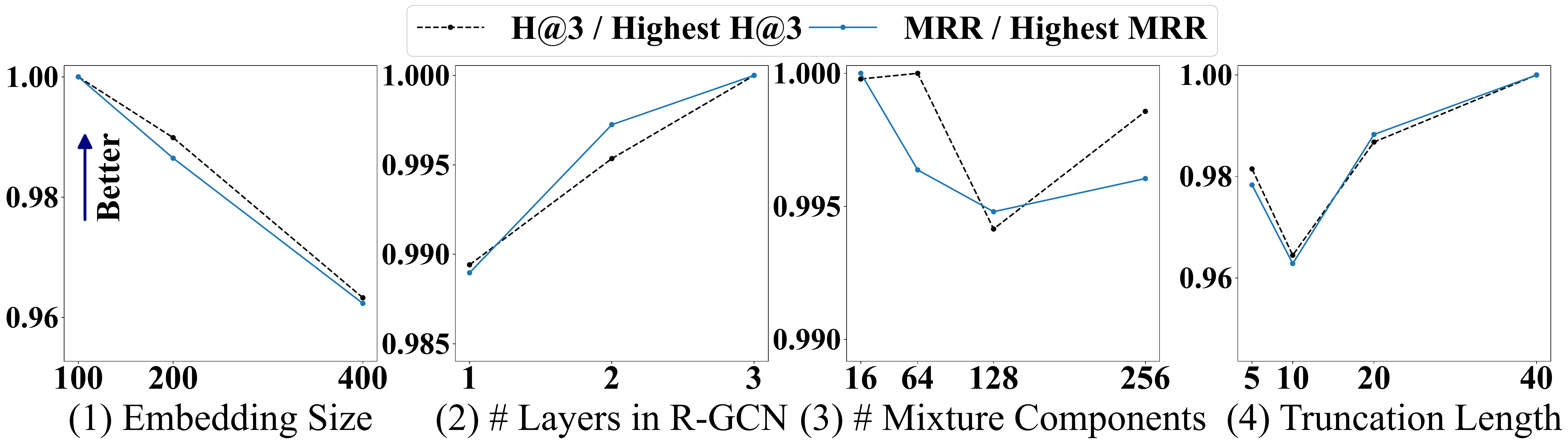}
		\setlength{\abovecaptionskip}{-1.0em}
		\caption{Temporal link prediction performance of \method on \icews.}
		\label{fig:exp:ablation:link}
	\end{subfigure}
}
\makebox[0.4\textwidth][c]{
	\begin{subfigure}[b]{0.55\textwidth}
		\centering
		\includegraphics[width=1.0\textwidth]{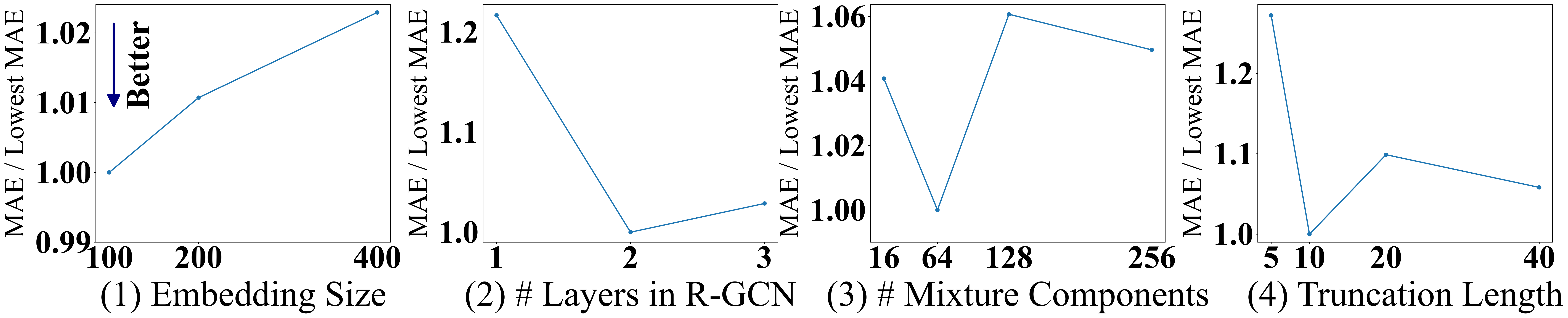}
		\setlength{\abovecaptionskip}{-0.9em}
		\caption{Event time prediction performance of \method on \icewsSmall.}
		\label{fig:exp:ablation:time}
	\end{subfigure}
}
\setlength{\abovecaptionskip}{0.75em}
\captionsetup{width=1.10\linewidth}
\caption{Link prediction and event time prediction performance 
	as we vary (1) embedding size, (2) number of R-GCN layers, (3) number of mixture components, and (4) truncation length.}
\label{fig:exp:ablation}
\vspace{-0.5em}
\end{figure}

We evaluate how the performance of \method changes, as we vary 
(a) the embedding size, 
(b) the number of R-GCN layers, 
(c) the number of mixture components, and
(d) truncation length (the number of time steps between backpropagation truncation in RNNs).
\Cref{fig:exp:ablation} shows the link prediction result on \icews (top), and event time prediction result on \icewsSmall (bottom);
reported values denote the ratio of the result obtained with the parameter setting on the x-axis to the best result.

\textbf{Embedding Size.} 
We set the embedding size (both temporal and structural embeddings) to 100, 200, and 400.
As \Cref{fig:exp:ablation:link,fig:exp:ablation:time} show, 
the best accuracy on the two datasets is achieved by an embedding size of 100, 
while using a much larger embedding size of 400 hurts the performance, as this leads to overfitting.

\textbf{Number of R-GCN Layers.}
\method extends R-GCNs for learning temporal and structural representations.
The number of R-GCN layers determines the size of the neighborhood from which a node aggregates information.
For predicting both temporal link and event time, using two layers leads to a better result than using a single layer, 
indicating that an increased neighborhood brings useful information for modeling a TKG.
However, using more layers can incur over-smoothing issues, decreasing time prediction accuracy.

\textbf{Number of Mixture Components.}
\method uses a mixture distribution to model the event time.
The number of mixture components affects the flexibility of the mixture distribution.
\Cref{fig:exp:ablation:link,fig:exp:ablation:time} report the performance of \method 
as the number of mixtures is set to 16, 64, 128, and 256.
\method achieves the best link and event time prediction results, with 16 and 64 mixture components, respectively.
While using a larger number of mixture components decreases performance, 
\method still achieves high accuracy, and is not very sensitive to these parameter settings.

\textbf{Truncation Length.}
For efficient and scalable training, \method truncates backpropagation every $ b $ time steps (Algorithm~\ref{alg:learning}).
We set $ b $ to 5, 10, 20, and 40, and measure the performance.
On the two datasets, the best result is achieved with $ b=40 $ (link prediction) and $ b=10 $ (event time prediction), 
and using a smaller time steps tends to decrease the accuracy,
as this restricts the model's ability to keep track of the history.
Results also show that if the truncation length is longer than appropriate, it may hurt the predictive accuracy.

\subsubsection{Effects of Event Time Modeling}\label{sec:exp:ablation:timecontribution}

To evaluate the importance of modeling event time in the overall quality of TKG modeling,
we report in \Cref{fig:exp:time_modeling_effects} the improvement made by event time modeling in terms of link prediction accuracy on all TKGs.
Specifically, let $ \text{Acc}_{\text{1,2}} $ and $ \text{Acc}_{\text{2}} $ be the link prediction performance 
obtained with both terms and only the second term in \Cref{eq:quad_prob}, respectively.
The improvement in \Cref{fig:exp:time_modeling_effects} is defined to be 
$ ((\text{Acc}_{\text{1,2}} - \text{Acc}_{\text{2}}) / \text{Acc}_{\text{2}}) \times 100 $.
Results show that modeling event time consistently improves the prediction accuracy on all datasets, by up to 61\%.

%% file: 050related.tex
In this section, we review previous works on reasoning over graphs. 

\textbf{Reasoning over Static Graphs.}
Inspired by the success of the Skip-gram model~\cite{DBLP:conf/nips/MikolovSCCD13} in NLP, 
several methods~\cite{DBLP:conf/kdd/GroverL16,DBLP:conf/kdd/PerozziAS14,DBLP:conf/www/TangQWZYM15}
learn node embeddings that maximize the likelihood of preserving neighborhoods of nodes in a network via random walks.
More recently, many graph neural networks (GNNs) have been developed for representation learning in homogeneous graphs for semi-supervised and self-supervised settings, 
including GCN~\cite{DBLP:conf/iclr/KipfW17} and GAT~\cite{DBLP:conf/iclr/VelickovicCCRLB18}.

\begin{figure}[t!]
	\par\vspace{-0.5em}\par
	\centering
	\includegraphics[width=0.475\textwidth]{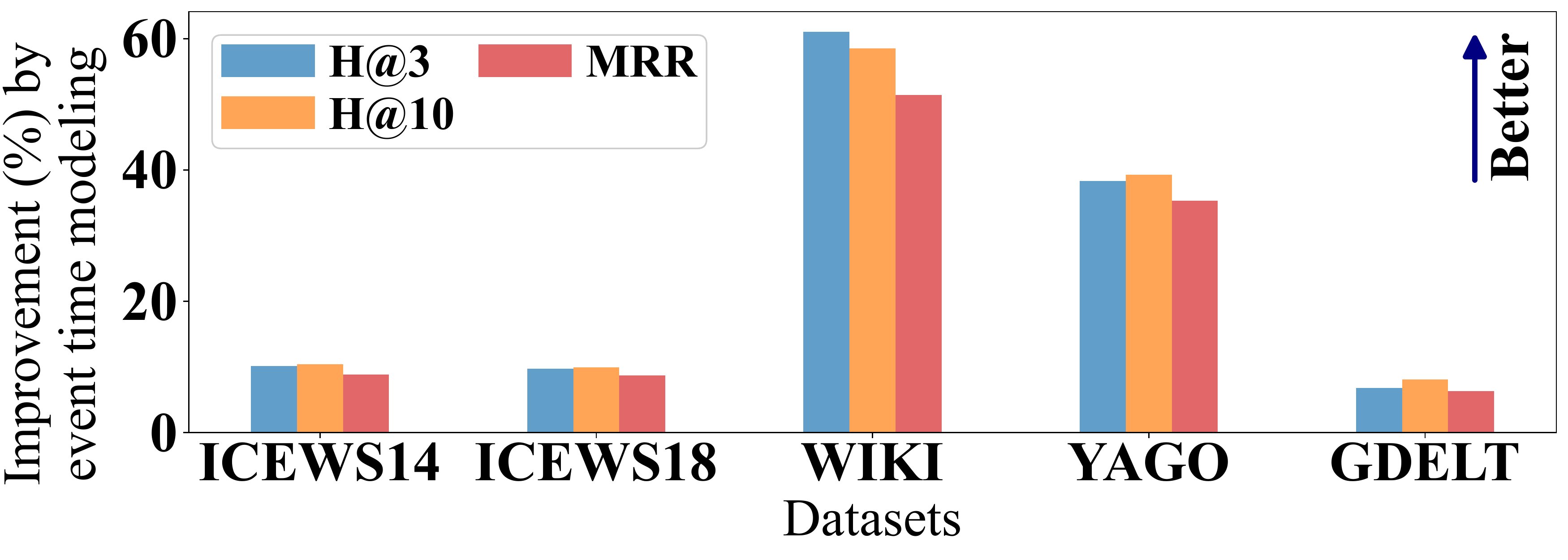}
	\setlength{\abovecaptionskip}{-1.5em}
\captionof{figure}{Modeling event time improves temporal link prediction accuracy on all TKGs, by up to 61\%.}
	\label{fig:exp:time_modeling_effects}
	\vspace{-2.0em}
\end{figure}

{To learn the representations of entities and relations in heterogeneous KGs, 
tensor factorization (TF)~\cite{DBLP:journals/siamrev/KoldaB09} has been widely used.
There exist several types of TF methods~\cite{DBLP:conf/cikm/ParkJLK16,DBLP:journals/vldb/ParkOK19,DBLP:journals/tpds/OhPJSK19,DBLP:conf/icde/JeonJSK16,DBLP:conf/www/GujralPP20}, 
such as CP and Tucker decomposition, 
which make different assumptions on the underlying data generating process.
Yet, most TF methods are not well suited for temporal data (e.g., they do not take inter arrival times into account).
In recent years, various relational learning techniques have been proposed for heterogeneous KGs,}
using different scoring functions to evaluate the triples in KGs,
including models with distance-based scoring functions (e.g., TransE~\cite{DBLP:conf/nips/BordesUGWY13}, RotatE~\cite{DBLP:conf/iclr/SunDNT19}) and
models based on semantic matching (e.g., RESCAL~\cite{DBLP:conf/icml/NickelTK11}, DistMult~\cite{DBLP:journals/corr/YangYHGD14a}, NTN~\cite{DBLP:conf/nips/SocherCMN13}, ConvE~\cite{DBLP:conf/aaai/DettmersMS018}).
GNNs have also been extended for relation-aware representation learning on KGs,
such as R-GCN~\cite{DBLP:conf/esws/SchlichtkrullKB18} and HAN~\cite{DBLP:conf/www/WangJSWYCY19}.
Overall, these methods are developed for static graphs and lack the ability to model temporally evolving dynamics.

\textbf{Reasoning over Dynamic Homogeneous Graphs.}
To capture temporal dynamics in time-evolving graphs, 
RNNs have been used to summarize and maintain evolving entity states in many methods~\cite{DBLP:conf/iconip/SeoDVB18,DBLP:journals/corr/abs-2006-10637,DBLP:conf/aaai/ParejaDCMSKKSL20,DBLP:conf/kdd/KumarZL19,DBLP:conf/ijcai/SingerGR19}.
Often, GNNs have been combined with RNNs 
to capture both structural and temporal dependencies~\cite{DBLP:conf/iconip/SeoDVB18,DBLP:journals/corr/abs-2006-10637,
DBLP:conf/aaai/ParejaDCMSKKSL20}.
Another line of work~\cite{DBLP:conf/iclr/XuRKKA20,DBLP:conf/wsdm/SankarWGZY20} 
employed graph attention mechanisms to make the model aware of the temporal order and the time span between entities when computing the attention weights.
Some other approaches applied deep autoencoders to dynamic graph snapshots~\cite{DBLP:journals/corr/abs-1805-11273,DBLP:journals/kbs/GoyalCC20},
enforced temporal smoothness on entity embeddings~\cite{DBLP:journals/tkde/ZhuGYSG16,DBLP:conf/aaai/ZhouYR0Z18}, and
performed temporal random walks~\cite{DBLP:conf/www/NguyenLRAKK18}.
As these methods are designed for single-relational dynamic graphs, they lack mechanisms to capture the multi-relational nature of TKGs,
which we focus on in this paper.

\textbf{Reasoning over Dynamic Heterogeneous Graphs.}
Static KG embedding methods have been extended to take temporal information into account,
including TA-DistMult~\cite{DBLP:conf/emnlp/Garcia-DuranDN18}, TTransE~\cite{DBLP:conf/www/LeblayC18}, HyTE~\cite{DBLP:conf/emnlp/DasguptaRT18}, and diachronic embedding~\cite{DBLP:conf/aaai/GoelKBP20}.
These temporal KG embedding techniques address an interpolation problem where the goal is to infer missing facts at some point in the past,
and cannot predict future events.
Recently, several methods have been developed to tackle the extrapolation problem setting, 
where the goal is to predict new facts at future time steps.
TensorCast~\cite{DBLP:conf/icdm/AraujoRF17} uses exponential smoothing to forecast latent entity representations, obtained with TF.
RE-Net~\cite{DBLP:conf/emnlp/JinQJR20} learns dynamic entity embeddings by summarizing concurrent events in an autoregressive architecture;
yet, it has no components to model the event time.
Know-Evolve~\cite{DBLP:conf/icml/TrivediDWS17}, DyRep~\cite{DBLP:conf/iclr/TrivediFBZ19}, and GHNN~\cite{DBLP:conf/akbc/HanM0GT20}
model the occurrences of events over time by using temporal point processes (e.g., Rayleigh and Hawkes processes)
that estimate the conditional intensity function.
Inspired by~\cite{DBLP:conf/iclr/ShchurBG20}, \method models the event time 
by directly estimating its conditional density in a flexible and efficient framework.

In summary, most existing methods for both homogeneous and heterogeneous dynamic graphs
model just the second term on the evolving network structure in~\Cref{eq:quad_prob}, 
and thus cannot predict when events will occur.
On the other hand, a few methods like \cite{DBLP:conf/icml/TrivediDWS17,DBLP:conf/iclr/TrivediFBZ19}
that model the first term on the evolving temporal patterns in~\Cref{eq:quad_prob}
do not model the other term, which greatly limits their reasoning capacity.
In this paper, we present a problem formulation that unifies these two major tasks (\Cref{sec:problem}), and 
develop an effective framework \method that tackles them simultaneously.

%% file: 060conclusion.tex
Temporal knowledge graphs (TKGs) represent facts about entities and their relations, 
which occurred at a specific time, or are valid for a specific duration of time.
Reasoning over TKGs, i.e., inferring new facts from TKGs, is crucial to many applications, including question answering and recommender systems.
Towards an effective reasoning over TKGs, this paper makes the following contributions.
\begin{itemize}[nosep,leftmargin=1em]
\item \textbf{Problem Formulation}.
We present a problem formulation 
that unifies the two core problems for TKG reasoning---modeling the timing of events and the evolving network structure.
\item \textbf{Framework}.
We develop \method, an effective framework for modeling TKGs 
that jointly addresses the two core problems.
\item \textbf{Effectiveness \& Efficiency}.
Experiments show that \method outperforms existing methods 
in terms of effectiveness (link and time prediction accuracy improved by up to \textit{116\%}) and 
efficiency (training speed improved by up to \textit{30$ \times $} over the best baseline).

\end{itemize}

\textbf{Reproducibility.} The code and data are available at \url{\codedataurl}.

\textbf{Future Work.}
We plan to improve the explainability and transparency of \method,
such that \method can answer questions like 
``When entity $ i $ is estimated to interact with entity $ j $, which past events had great influence on their current relationship?''.
We also plan to extend \method for anomaly detection in dynamic networks.

%% file: 070appendix.tex
\subsection{Experimental Settings}\label{appendix:settings}

\indent\indent
\textbf{Data Split.} 
We split datasets into training, validation, and test sets in chronological order, as shown in \Cref{tab:datasets}.
For training \method, we applied early stopping, checking the validation performance with a patience of five.
Then the model with the best validation performance was used for testing.

\textbf{Hyperparameters.}
We used a two-layer R-GCN~\cite{DBLP:conf/esws/SchlichtkrullKB18} with block diagonal decomposition (BDD),
which reduces the number of parameters and  alleviates overfitting, and
set the size of entity and relation embeddings in \method to 200 
(except for \wiki, where it was set to 192 to meet the constraint of using R-GCN with BDD).
Static entity embeddings were initialized using the Glorot initialization,
while the initial dynamic embeddings were zero-initialized.
We used a single-layer Elman RNN with tanh non-linearity, but different RNNs, such as GRU, can easily be used.
We trained the model using the AdamW optimizer 
with a learning rate of 0.001, a weight decay of 0.00001, $ \beta_1=0.9 $, and $ \beta_2=0.999 $, 
and applied dropout with $ p=0.2 $.
As training the module for modeling network structure usually takes longer than training the module for modeling event time,
we first trained the model with $ \lambda_1\!=\!0$ and $ \lambda_2\!=\!1 $, and 
then trained the entire model with $ \lambda_1\!=\!\lambda_2=1 $ until convergence.
We truncated the backpropagation for RNNs every 40 time steps for \gdelt, and every 20 time steps for other datasets.
We set the number $ K $ of mixture components to 128.

For the details of baselines used in this work, please refer to 
\cite{DBLP:conf/icml/TrivediDWS17} for \knowevolve, \rtpp, and \mhp;
\cite{DBLP:conf/akbc/HanM0GT20} for \ghnn and \litsee; and
\cite{DBLP:conf/emnlp/JinQJR20} for other baselines including \renet.

\textbf{Compute Resources.}
We ran experiments on a Linux machine with 8 CPUs (Intel(R) Xeon(R) CPU E5-2623 v4 @ 2.60GHz), 30GB RAM, and an NVIDIA Quadro P6000 GPU.

\textbf{Software.}
To implement \method and the evaluation pipeline, we used the following software (software version is specified in the parentheses):
python (3.8.3), Deep Graph Library (0.53), PyTorch (1.7.1), NumPy (1.18.5), and pandas (1.0.5).
We used PyTorch's RNN implementation, and Deep Graph Library's R-GCN implementation.